\documentclass{article}

\usepackage{graphicx} % Required for inserting images
\usepackage{booktabs}
\usepackage{multirow}
\usepackage{}

\usepackage{subfig}

\usepackage{changepage,threeparttable}

\usepackage{tcolorbox}
\usepackage{float}
\usepackage{subfloat}
\usepackage{subfig}

\usepackage{booktabs}
\usepackage{amsmath}
\usepackage{multirow}
\usepackage{multicol}
\usepackage[round]{natbib}

\usepackage[final]{acl}

\usepackage{times}
\usepackage{latexsym}
\usepackage{amsfonts}
\usepackage{algorithm2e}
\usepackage[T1]{fontenc}
\usepackage[utf8]{inputenc}

\usepackage{microtype}

\usepackage{inconsolata}

\title{NLP needs Diversity outside of `Diversity'}

\author{
Joshua Tint\thanks{Paper was written prior to employment by Amazon} \\
  Arizona State University\\ 
  \texttt{jrtint@asu.edu} 
}
\date{February 2025}

\begin{document}

\maketitle

%\author{
%  \textbf{First Author\textsuperscript{1}},
%  \textbf{Second Author\textsuperscript{1,2}},
%  \textbf{Third T. Author\textsuperscript{1}},
%  \textbf{Fourth Author\textsuperscript{1}},
%\\
%  \textbf{Fifth Author\textsuperscript{1,2}},
%  \textbf{Sixth Author\textsuperscript{1}},
%  \textbf{Seventh Author\textsuperscript{1}},
%  \textbf{Eighth Author \textsuperscript{1,2,3,4}},
%\\
%  \textbf{Ninth Author\textsuperscript{1}},
%  \textbf{Tenth Author\textsuperscript{1}},
%  \textbf{Eleventh E. Author\textsuperscript{1,2,3,4,5}},
%  \textbf{Twelfth Author\textsuperscript{1}},
%\\
%  \textbf{Thirteenth Author\textsuperscript{3}},
%  \textbf{Fourteenth F. Author\textsuperscript{2,4}},
%  \textbf{Fifteenth Author\textsuperscript{1}},
%  \textbf{Sixteenth Author\textsuperscript{1}},
%\\
%  \textbf{Seventeenth S. Author\textsuperscript{4,5}},
%  \textbf{Eighteenth Author\textsuperscript{3,4}},
%  \textbf{Nineteenth N. Author\textsuperscript{2,5}},
%  \textbf{Twentieth Author\textsuperscript{1}}
%\\
%\\
%  \textsuperscript{1}Affiliation 1,
%  \textsuperscript{2}Affiliation 2,
%  \textsuperscript{3}Affiliation 3,
%  \textsuperscript{4}Affiliation 4,
%  \textsuperscript{5}Affiliation 5
%\\
%  \small{
%    \textbf{Correspondence:} \href{mailto:email@domain}{email@domain}
%  }
%}

\begin{abstract}
    This position paper argues that recent progress with diversity in NLP is disproportionately concentrated on a small number of areas surrounding fairness. We further argue that this is the result of a number of incentives, biases, and barriers which come together to disenfranchise marginalized researchers in non-fairness fields, or to move them into fairness-related fields. We substantiate our claims with an investigation into the demographics of NLP researchers by subfield, using our research to support a number of recommendations for ensuring that all areas within NLP can become more inclusive and equitable. In particular, we highlight the  importance of breaking down feedback loops that reinforce disparities, and the need to address geographical and linguistic barriers that hinder participation in NLP research. 
\end{abstract}

\section{Introduction}

Natural language processing, like many subfields of computer science, has long been male-dominated, and subject to severe WEIRD (Western, Educated, Industrialized, Rich, and Developed) biases \cite{hovy2021five, mihalcea2024ai}. As a science studying the fundamentally social phenomenon of language, it is imperative that the researchers working in NLP reflect the world it serves. Recently, the diversity of the field of NLP has been reported by many sources to be increasing, including more women and people from varied geographic backgrounds \cite{muscato-etal-2024-overview, khanuja-etal-2023-evaluating}. In the last five years, the ACL has introduced multiple initiatives aimed at improving the fairness and inclusivity of the field, including adding an new Diversity \& Inclusion committee in 2020 \cite{ACL2020DI}. However, we find that these gains are in fact not as vast as they appear—that diversity in NLP is largely concentrated in areas such as fairness and ethics, and have largely not penetrated ``core'' NLP fields. We perform novel research to analyze the current state of diversity in various topics within NLP, and speculate as to why many areas have been left behind in terms of inclusion. We synthesize our research to create a set of recommendations for both improving the true diversity of NLP, and for accurately reporting diversity statistics.

\subsection{What is ``Diversity,'' anyways?}

We take from Cletus a definition of diversity as ``mutual respect for qualities and experiences of individuals who have different
attributes'' \cite{cletus2018prospects}. The attributes in question are not limited in scope, and may include social class, race, orientation, religion, nationality, disability, and ancestry \cite{hattery2022diversity}. Some categories, such as caste, are social frameworks that exist within only a few cultures \cite{deshpande2015caste}. For the sake of clarity, we focus on two categories, where a bias is widely observed and easily understood: gender and geography. Following Devinney's recommendations, this paper will present a theory of gender in the interest of precision \cite{theories-of-gender}. This paper recognizes the existence of nonbinary genders outside male and female, and furthermore that gender and biological sex are distinct. However, we do examine datasets which do not provide data on nonbinary researchers due to a lack of available data. %These have the benefit of being widely reported and easily obtainable pieces of information 

\subsection{Diversity in NLP}

Various attempts have been made to quantify the gender diversity of the field at different points in time. Overall, NLP researchers are far more likely to be male than female. In 2020, Mohammad performed a comprehensive study of gender bias in authorship of NLP papers, estimating that approximately 29\% of authors were female \cite{mohammad-2020-gender}. Unfortunately, while Mohammed provides data on gender bias by paper topic, fairness-related topics were not included in that analysis. The ACL publishes membership statistics, but the most recent dataset is as of 2017; in that year, 24\% of members were female (while 70\% were male and 6\% were unknown). 

In terms of geographic diversity, there is less data available. NLP researchers are more concentrated in Europe and North America, while less concentrated in other regions. From the ACL's 2017 dataset, 55\% of authors were from the Americas, 19\% were from Europe, Africa or the Middle East, and 23\% were from Asia or Oceania (3\% were unknown) \cite{acl-diversity-data}. % ADD MORE

\section{NLP's Diversity by Topic}

Facing a gap in the research, we were motivated to perform a small experiment to understand how  diversity in NLP varies by topic. 

\subsection{Methodology}

In order to understand the composition of researchers in a given field, we use a keyword search on \texttt{ACL Anthology} to identify authors that are prolific in a particular area. For each keyword, we performed an author page search using the default ``Relevance'' ranking algorithm. ACL uses Google’s Programmable Search algorithm to prioritize pages that recently and prevalently use the keyword, or text semantically similar to the keyword \cite{howsearchwork}. We filtered by author pages to identify authors that are prolifically and recently associated with these keywords (i.e. authors that publish often in the field and are still active). For each keyword, we identify the top 50 authors using the aforementioned ``Relevance'' ranking, identifying the author's gender and geographical location. 

To find the author's gender, we first check their ACL profile, then if a gender is unavailable we perform a web search to identify any profiles on a personal website or from affiliated institutions. If no gender is found, we mark the gender as ``unknown,'' but this was only needed in one instance. We took use of traditionally gendered pronouns, e.g. ``he'' or ``she'' as an indicator of male and female, respectively, following the method used by \citet{stathoulopoulos2019gender}. Pronouns such as these were often found when researchers were discussed in works by outside sources, or when researchers marked pronoun declensions in their personal biographies or profiles (e.g. ``FirstName LastName, she/her''). Due to methodological and social concerns \cite{gautam-etal-2024-stop}, we do not use names as an indicator of gender, unlike much prior research on gender disparities in NLP \cite{ding2023voicesheranalyzinggender}. Resources that were used to determine authors' gender include the following. Ordering indicates the precedence in which we used each resource:

\begin{enumerate}
    \item Pages about authors from their affiliated institutions, e.g. biographies or articles written about them in campus news
    \item Pages about authors from other sources, such as news articles
    \item Authors' personal websites or online CVs 
    \item Authors' professional profiles on career-related or blogging platforms, e.g. LinkedIn or X.com (formerly Twitter)
\end{enumerate}

To identify keywords pertaining to NLP fields, we use the keyword extraction performed by \citet{mohammad-2020-gender} in NLP papers . Of the top 20 keywords identified, we for filter those which are associated with a field or topic of research (for instance, eliminating ``large scale''). This leaves us with 12 remaining keywords, which can be seen in Table \ref{tab:keywords}. However, this analysis had no fairness related keywords, thus we use the three identified in the ACL's fairnes track: ``fairness,'' ``bias,'' and ``ethics.''   This approach was chosen over using the ACL's tracks, as many tracks make poor keywords for searchability, e.g., work in NLP applications rarely contains the phrase ``NLP Applications.'' ARR topic area keywords were considered as well, but most are too narrow to represent a significant portion of the NLP community. Of the keywords we used, there was some potential for overlap, for instance between ``Language Models'' and ``Word Embeddings,'' however, no authors appeared under more than one query. We included 150 distinct authors in fairness-related categories, and 600 in non-fairness related categories.

\begin{table}[!htp]\centering
\caption{Topic-based keywords used in measuring diversity}\label{tab:keywords}
\scriptsize
\begin{tabular}{lp{4cm}}\toprule
Non-fairness-related &Language Models, Machine Translation, Neural Networks, Optimization, Sentiment Analysis, Question Answering, Entity Recognition, Speech Recognition, Segmentation, Relation Extraction, Word Embeddings, Semantic Parsing \\\midrule
Fairness-related &Bias, Fairness, Ethics \\
\bottomrule
\end{tabular}
\end{table}

As a proxy for understanding the author's geographical location, we marked the location of the author's current affiliated institution, rather than birth place or nationality. This is because the later characteristics are much more difficult to obtain, and because it provides a less complete picture of where researchers are currently performing work. A university's geography can strongly influence what labs and projects are available there, and can shape research directions through funding as well \cite{radinger2019influences}.  We recorded the information at the continent level, with the categories being: ``North America,'' ``South America,'' ``Europe,'' ``Asia,'' ``Africa,'' and ``Oceania,'' following the borders of the United Nations Geoscheme \cite{UNGeoscheme}.  This binning was chosen over the ACL's given that it only records geography by three categories: Americas, Europe/Africa/Middle East, and Asia Pacific. This categorization is not particularly useful for fairness research (e.g. because the under-represented South America is binned with the over-represented North America, and so on). Again, we first try to identify this information from the authors' ACL profiles, then perform a web search for profiles on a personal website or from affiliated institutions. Again, we provided for the possibility of marking a continent as ``unknown'' but this marking was never needed.

\subsection{Findings}

\begin{figure*}[ht]
    \centering
    \subfloat[]{
        \includegraphics[width=0.49\textwidth]{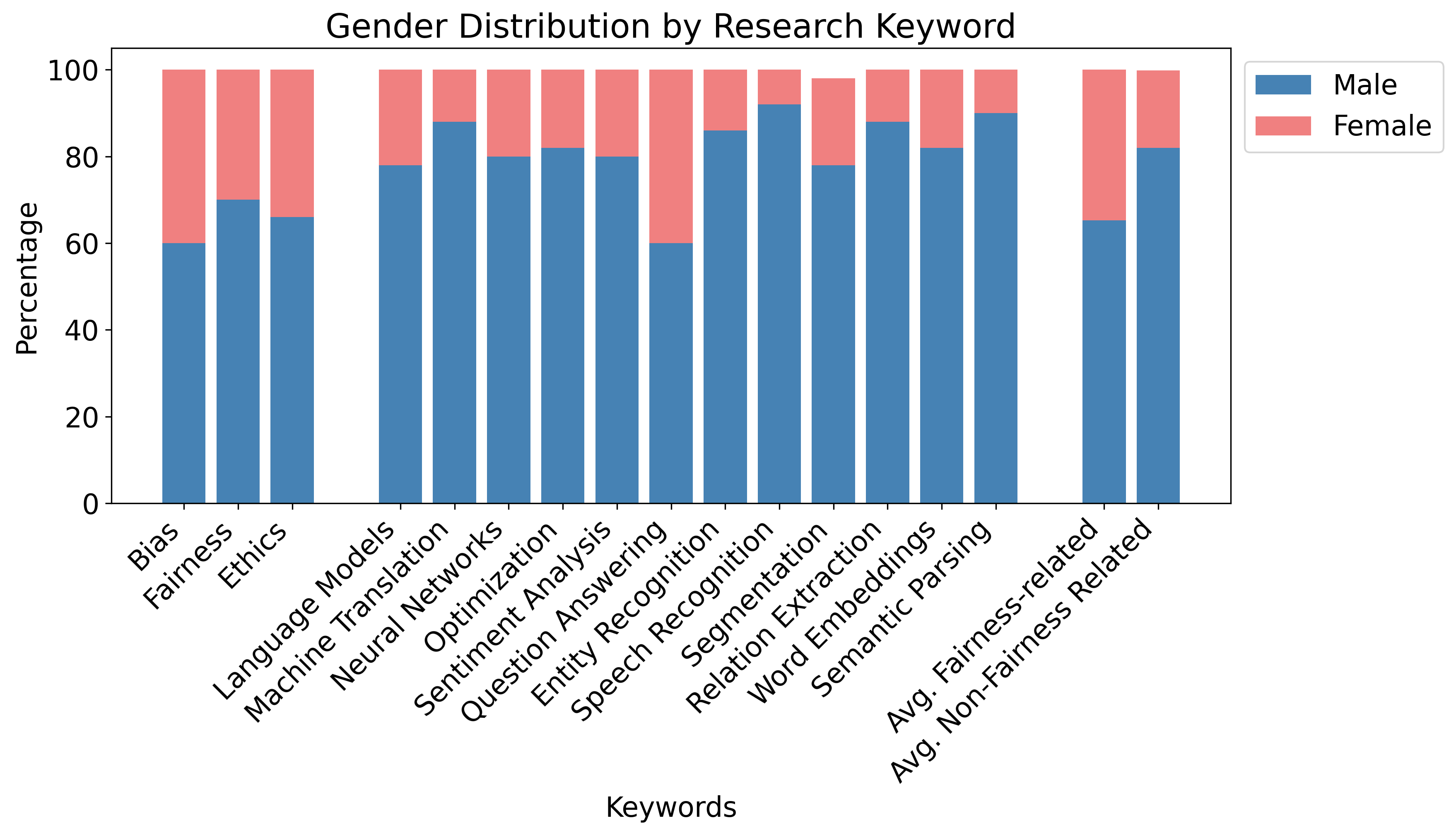}
    }
    \subfloat[]{
        \includegraphics[width=0.49\textwidth]{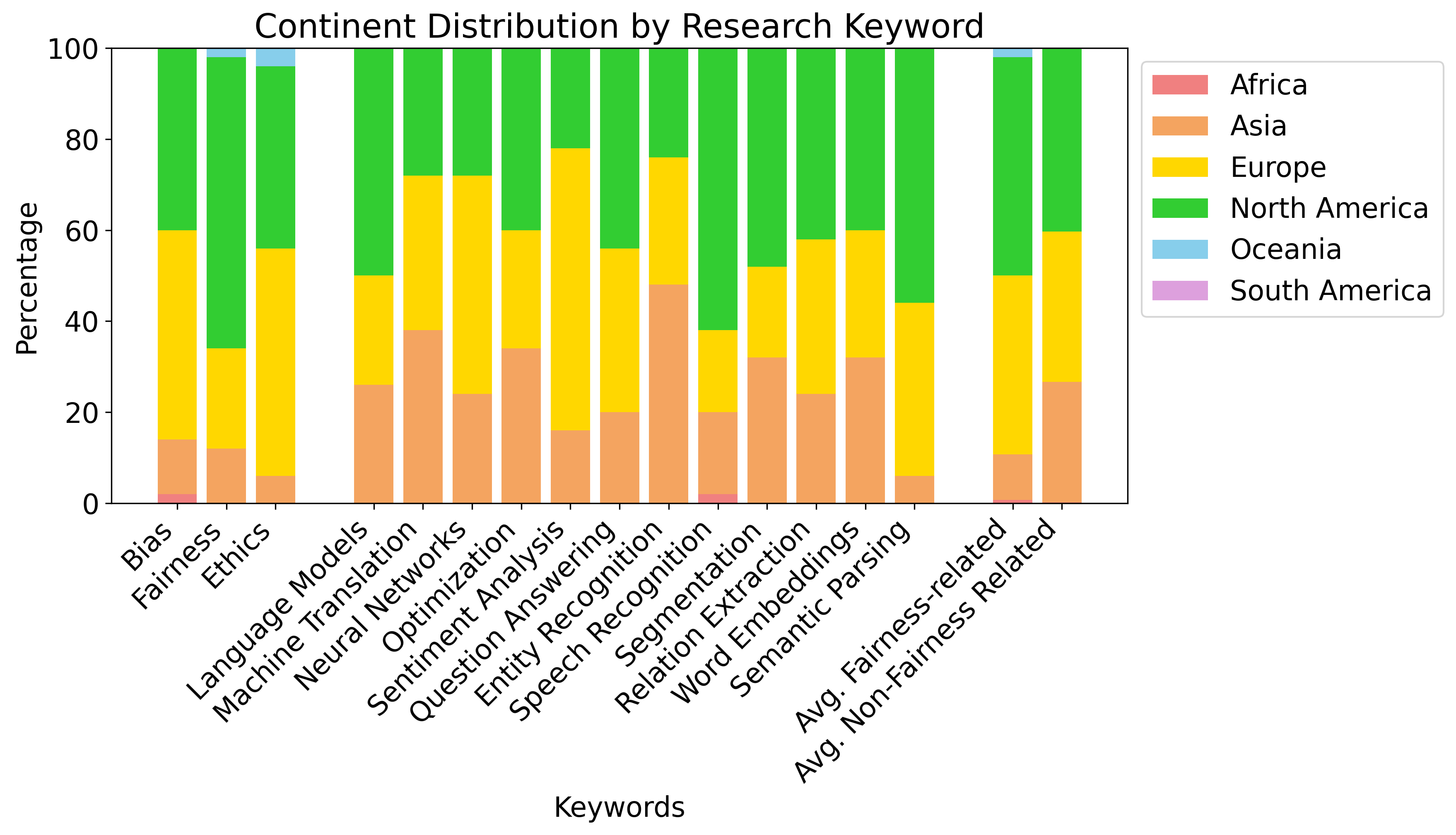}
    }

    \caption{The gender (a) and continents (b) of top researchers by research keyword}
    \label{fig:gender}
\end{figure*}

Collating the gender of the researchers in each keyword, the differences between fairness- and non fairness-related fields is stark. For fairness-related keywords, the average percentage of women was 34\%, while for non-fairness related keywords the average was 18\%, almost half. While neither field has gender parity, fairness-related fields clearly have gender diversity that far eclipses non-fairness related ones. Overall, our results largely comport with the ACL's overall gender membership data, which reported 24\% women. The complete results for gender data can be found in Table \ref{tab:genders}.

Collecting the continental affiliations of researchers, it's clear that there is also a strong divide between those in fairness-related and non-fairness related fields. North America is more strongly represented in fairness-related fields (48\% instead of 40\%), while Asia is less strongly represented (10\% instead of 27\%). The complete results for continent data can be found in Table \ref{tab:continent}.

Analyzing the geography of researchers is somewhat more tricky. For gender, it is widely accepted that the ideal outcome is to have as many male as female researchers (gender parity), but the ideal state of geographic diversity is less clear. It has been argued for a more even balance of representation by region \cite{knutsen2010question}, for a distribution which mirrors the population of each region \cite{skupien2020geography}, or for a more sophisticated distribution which, for instance, reflects the number of universities in each region \cite{fitzgerald2021academia}. The exact proportion that each continent should occupy is highly arguable. One could argue that by having a lesser proportion of researchers affiliated with Asian institutions, the fairness-related fields are in some ways less diverse, and that argument has merit. However, the fairness-related fields have more continents represented in total. It's also remarkable that all Oceanian researchers and one of the only two African researchers in the dataset appeared under a fairness-related keyword, particularly because the fairness group contained one quarter of the total authors. Taking this into account, it's clear that while fairness-related areas are by no means perfectly diverse, that non-fairness related fields may have a severe issue with geographic diversity.

\section{Explaining the Diversity Gap}

On its own, a gap in diversity in fairness-related fields is unsurprising. As a form of self-selection bias, it is understandable that those impacted the most by the lack of diversity in NLP would be disproportionately predisposed to join topics in fairness. However, we believe that there is substantial evidence of structural issues that pervade NLP which deepen this divide.  

Cultural and institutional climates of different subfields may contribute to the observed disparities. Studies have shown that women and researchers from marginalized communities often face higher barriers in STEM, including biases in work distribution and collaboration opportunities \cite{doi:10.1126/sciadv.1400005, witteman2019gender}. Fairness-related fields may provide a comparatively more inclusive environment due to the nature of the research itself, which prioritizes diversity, equity, and inclusion. This could lead to a feedback loop where underrepresented researchers feel more welcome and supported in fairness-related subfields, reinforcing the diversity gap in other areas.

The presence of role models is a key factor in guiding students into research areas \cite{gibson2004role}. In particular, students may look to role models of their gender or specific minority group as influences in their decisions \cite{nelson2003national}. It is possible that on top of self-selection bias, that incoming NLP students are pushed towards fairness by the presence of more like role models in the field already. In this sense, the diversity in fairness is self-sustaining, while the relative lack of diversity in other fields may be a vicious cycle. Additionally, faculty in a field may express homophily, a preference (implicit or explicit) towards potential students in similar groups as them \cite{milkman2015happens, moss2012science}. For instance, a male professor may favor male graduate students. This narrows the entry pathways for underrepresented future researchers in fields with lower diversity.

Funding and hiring structures could also help create this disparity. ``Big Tech'' companies such as Google, Amazon, and Microsoft are common sponsors and sources of funding for machine learning research, increasingly so in recent years \cite{birhane2022valuesencodedmachinelearning, attema2025public}. Work sponsored by these companies is more likely to contain ethical statements and discuss potential negative outcomes, such as negative fairness outcomes \cite{birhane2022valuesencodedmachinelearning}. Meanwhile, Ding finds that top female researchers are more prominent at these large tech companies versus universities, which they attribute to a difference in hiring standards \cite{ding2023voicesheranalyzinggender}. Taken together, this may create an environment where female researchers (and possibly researchers from other underrepresented categories) are more incentivized to join fairness research. However, more direct research on this topic would be needed to definitively prove such a proposition.

There are also fundamental technological issues which prevent some researchers from properly engaging with many ``core NLP'' areas. Linguistic diversity in language technologies is a greatly-studied and pervasive issue; as few as 7 languages are widely supported across datasets and models, with English dominating by a large margin \cite{joshi-etal-2020-state, samardzic2024measure}. Without models or datasets that support a researcher's native language, many downstream research tasks become close to impossible. This could \textit{de facto} move many researchers into the fairness-adjacent area of linguistic diversity.

\section{Recommendations}

We make a set of recommendations from our results. They highlight where progress has been made and where further action is needed to ensure diversity is not confined to fairness-focused research areas.

\textbf{1. Improve Diversity Statistics:} There is an unjustifiable lack of transparency in the ACL's own diversity reporting data. Its most recent public report is from 2017 and does no crosstabulation by research field, topic, or track. Data from other researchers is often more recent, but less comprehensive, and does not take into account the discrepancy between fairness and non-fairness areas. Reporting information about the diversity of subfields is a necessary step for ensuring that NLP is diverse not just superficially but at all levels. This transparency could highlight other areas which are also substantially more (or less) diverse than average, which would be necessary to stimulate any further change to rectify these issues. Additionally, 

\textbf{2. Break Down Feedback Loops:} Both the issue of role models and the issue of cultural climates present self-perpetuating cycles that can keep marginalized researchers out of non-fairness related fields. Promoting women and underrepresented groups who do important work in non-fairness NLP areas is important to show young researchers the breadth of paths they can take. Featuring these sorts of role models in conference keynotes, panels, or networking events—even as far as establishing explicit mentorship programs—could go a long way to making non-fairness related fields more equitable. These kinds of programs have been repeatedly recommended by those studying diversity in computational linguistics, but are much more widely adopted in fairness-related labs and organizations. To that end, non-fairness related groups should explore practices that are more commonplace in NLP fairness, like citation diversity statements \cite{zurn2020citation}, which could improve cultural climate. Additionally,  institutions could proactively highlight and support underrepresented researchers working in other domains to offer broader role models. The current lack of diversity in NLP is an enormous problem that has taken decades of progress to even get as far as it has, so these solutions on their own will likely not ``solve'' the problem, but they should form a solid foundation to build towards true fairness in all subfields.

\textbf{3. Address Geographical and Linguistic Barriers:} To foster more inclusivity in ``core NLP'' areas, it's essential to actively address geographical and linguistic barriers. Institutions and funding bodies should support initiatives aimed at broadening language representation in NLP research by creating more open-source datasets for underrepresented languages, particularly those spoken by marginalized communities. Encouraging the development of multilingual models and prioritizing linguistic diversity in training data can also empower researchers working in lower-resource languages. Once again, these are initiatives that have been recommended by others. Yet they often remain tied to fairness-related theory work (e.g., bias mitigation, ethical modeling), and face barriers entering practice in terms of generating new models and datasets \cite{10.1145/3351095.3372826}, which could dramatically lower barriers to entry into ``core NLP.''  Creating fellowship programs or grants specifically targeting researchers from diverse linguistic and geographic backgrounds can further ensure equitable access to resources and opportunities in NLP. While fellowship programs and global partnerships are emerging \cite{jumi2024multilingual}. they are often short-term, underfunded, or confined to specific regions or linguistic domains. Moreover, such initiatives frequently lack integration with major NLP conferences and research pipelines, limiting their impact on the composition of authors in core subfields. Without structural changes to publication incentives, reviewer expectations, and infrastructure access, researchers working outside the dominant academic centers may continue to be relegated to the periphery. A sustained effort is needed to embed linguistic and geographic inclusivity into the foundational research practices of NLP, rather than treating it as a side concern of fairness-oriented work.

\section{Conclusion}

The diversity gap between fairness and non-fairness fields in NLP is, in one sense, a failure—marginalized researchers have been pushed out of some areas and concentrated in others. But it is also a success story: fairness has become a subfield where diversity thrives. This likely reflects both exclusion from other areas and the welcoming climate within fairness work. Having diverse researchers study diversity is not inherently an issue, nor is the uneven distribution of diversity necessarily problematic. However, diversity is equally critical in non-fairness subfields, not only as an ethical imperative but because it demonstrably improves research quality, with mixed-gender authorship often outperforming single-gender teams \cite{gender-diversity}. Blanket-level diversity statistics obscure these imbalances, and true inclusion requires deeper reflection on where inclusivity has taken hold—and where it still has not.

\section{Limitations}

While the methodology presented offers valuable insights into the demographic composition of researchers in NLP, some limitations on our data collection for NLP researchers should be noted. First, the gender identification process, though using a common approach, is limited by the availability and accuracy of publicly accessible profiles, which may result in misclassifications. For instance, some authors may use pronouns that are different than is traditional for their gender. Additionally, if authors present a different gender at work than they do personally, our reliance on professional profiles and sources could lead to their gender being misclassified. Another concern is that the ACL Anthology as a source may not adequately link trans authors who have published under different names, potentially excluding them from analyses of prolific authorship and thereby erasing their contributions. We only present data in aggregate, reducing the potential harm of misgendering, these issues could lead nonbinary or transgender authors to be overlooked.

Geographical data, while informative, is based on current institutional affiliations, which may not fully represent the global distribution of researchers, as many authors may move between institutions or work in regions not captured by their current affiliations. This limitation is particularly relevant for researchers in less-documented regions, such as Africa or Oceania, where the availability of data may be sparse. Finally, the geographic categorization used in this study may oversimplify regional diversity by grouping large and culturally distinct regions together. This could obscure nuanced regional imbalances and create an incomplete picture of global diversity in NLP research.

\bibliography{bibliography}

\section{Appendix}

\subsection{Supplemental Tables}

Complete data for researcher gender can be found in Table \ref{tab:genders} and researcher continent can be found in Table \ref{tab:continent}.

\begin{table*}[!htp]\centering
\caption{Continent of researchers' affiliated institution, by researcher keyword}\label{tab:continent}
\scriptsize
\begin{tabular}{lrrrrrrr}\toprule
\multirow{2}{*}{Research Keyword} &Researcher Continent & & & & & \\
&Africa (\%) &Asia (\%) &Europe (\%) &North America (\%) &Oceania (\%) &South America (\%) \\\midrule
Bias &2&12&46&40&0&0\\
Fairness &0&12&22&64&2&0\\
Ethics &0&6&50&40&4&0\\
Language Models &0&26&24&50&0&0\\
Machine Translation &0&38&34&28&0&0\\
Neural Networks &0&24&48&28&0&0\\
Optimization &0&34&26&40&0&0\\
Sentiment Analysis &0&16&62&22&0&0\\
Question Answering &0&20&36&44&0&0\\
Entity Recognition &0&48&28&24&0&0\\
Speech Recognition &2&18 &18&62&0&0\\
Segmentation &0&32&20&48&0&0\\
Relation Extraction &0&24&34&42&0&0\\
Word Embeddings &0&32&28&40&0&0\\
Semantic Parsing &0&6&38&56&0&0\\
Avg. Fairness-related &0.7 &10&39.3 &48&2&0\\
Avg. Non-Fairness Related &0.2 &26.5 &33&40.3 &0&0\\
\bottomrule
\end{tabular}
\end{table*}

\begin{table*}[!htp]\centering
\caption{Gender of researcher, by researcher keyword}\label{tab:genders}
\scriptsize
\begin{tabular}{lrrrr}\toprule
\multirow{2}{*}{Research Keyword} &Researcher Gender & & \\
&Male (\%) &Female (\%) &Unknown (\%) \\\midrule
Bias &60&40&0 \\
Fairness &70&30&0 \\
Ethics &66&34&0 \\
Language Models &78&22&0 \\
Machine Translation &88&12&0 \\
Neural Networks &80&20&0 \\
Optimization &82&18&0 \\
Sentiment Analysis &80&20&0 \\
Question Answering &60&40&0 \\
Entity Recognition &86&14&0 \\
Speech Recognition &92&8&0 \\
Segmentation &78&20&2 \\
Relation Extraction &88&12&0 \\
Word Embeddings &82&18&0 \\
Semantic Parsing &90&10&0 \\
Avg. Fairness-related &65.3 &34.7 &0\\
Avg. Non-Fairness Related &82&17.8 &0.2 \\
\bottomrule
\end{tabular}
\end{table*}

\end{document}